# UniD³: A Knowledge Graph–Enhanced RAG Framework for Drug–Disease Discovery and Reasoning

Qing Wang, MS[1], Tianshi Liu, MD[2], Minghao Zhou, MS[1], Jialu Liang, MS[1], Sen Guo, PhD[1], Guangyu Wang, PhD[3,4], Jing Su, PhD[5], Qianqian Song, PhD[1,✉].

[1]Department of Health Outcomes and Biomedical Informatics, University of Florida, Florida, USA

[2]Department of Hematology, H. Lee Moffitt Cancer Center and Research Institute, Florida, USA

[3]Center for Bioinformatics and Computational Biology, Houston Methodist Research Institute, Houston, TX, USA

[4]Department of Cardiothoracic Surgery, Weill Cornell Medicine, Cornell University, New York, NY, USA

[5]Department of Biostatistics and Health Data Science, Indiana University School of Medicine, IN, USA

✉Corresponding authors: Qianqian Song, PhD: qsong1@ufl.edu.

## ABSTRACT

**Objectives:** Systematic characterization of drug–disease relationships is essential for drug discovery and repurposing, yet remains hindered by the heterogeneity and rapid growth of biomedical literature. Existing datasets rely on labor-intensive curation and are often incomplete, while LLM-only approaches suffer from hallucination and weak evidence grounding. We introduce UniD³, a unified framework that integrates Large Language Models with Knowledge Graph–enhanced Retrieval-Augmented Generation (KG-RAG) to systematically extract, organize, and validate drug–disease knowledge across Drug–Disease Matching (DDM), Drug Effectiveness Assessment (DEA), and Drug–Target Analysis (DTA) tasks.

**Methods:** UniD³ processes 157,849 PubMed articles using task-specific prompts with Llama 3.3-70B for classification. A dual-stage entity extraction strategy constructs knowledge graphs, by combining paper-level extraction with knowledge graph–level consolidation centered on drug and disease entities. These graphs support KG-RAG–based generation of structured datasets for DDM, DEA, and DTA. Dataset quality is evaluated through external benchmarks, fuzzy matching with curated resources, and clinician review. Multiple LLMs are benchmarked, and an explainable chatbot is developed for evidence-grounded question answering.

**Results:** UniD³ generates six knowledge graphs and large-scale datasets, including 28,915 DDM samples, 15,042 DEA samples, and over 4,000 DTA question–answer pairs. External validation demonstrates strong performance (F1: 0.85–0.87 for DDM/DEA; 0.82 for DTA), with clinician review confirming high reliability (AUROC = 0.90). KG-RAG–augmented models outperform standalone LLMs. The UniD³ chatbot enables interpretable, citation-supported exploration of drug–disease relationships and mechanisms.

**Conclusion:** UniD³ provides a scalable, extensible framework for transforming unstructured biomedical literature into high-quality, structured drug–disease knowledge, supporting AI-driven discovery, repurposing, and precision medicine.

## INTRODUCTION

Systematic characterization of drug-disease relationships is fundamental to biomedical research, underpinning drug discovery, repurposing, and precision medicine. Central to this endeavor are three tightly coupled tasks: Drug-Disease Matching (DDM)[1], which identifies therapeutic indications for pharmacological compounds; Drug Effectiveness Assessment (DEA), which evaluates drug performance across diseases, patient populations, and clinical contexts[2-6]; and Drug-Target Analysis (DTA), which elucidates the molecular mechanisms through which drugs exert their effects[7-9]. Together, these tasks link compounds, diseases, clinical outcomes, and biological targets, enabling mechanism-aware hypothesis generation, rational therapeutic selection, and data-driven treatment optimization[10]. Consequently, comprehensive Drug-Disease Datasets ($D^3$) that jointly support DDM, DEA, and DTA represent critical research infrastructure. However, constructing such datasets remains challenging due to the scale, heterogeneity, and rapid evolution of biomedical knowledge distributed across diverse textual sources.

Existing Drug-Disease datasets are predominantly built through manual curation or rule-based extraction pipelines[11]. While these approaches can achieve high precision, they are labor-intensive, difficult to scale, and slow to incorporate newly published evidence. As a result, many curated resources offer incomplete coverage, focus on narrow tasks, or become outdated as biomedical literature continues to expand[12, 13]. Recent advances in Large Language Models (LLMs) have opened new possibilities for automating the extraction of structured knowledge from unstructured biomedical text. Models such as the GPT[14] series and LLaMA[15] series demonstrate strong capabilities in biomedical language understanding, contextual reasoning, and large-scale information synthesis. These strengths have motivated their application to dataset construction in diverse settings, including data augmentation in low-resource environments[16], dialogue-driven structured information extraction[17], and clinically validated annotation of drug toxicity from regulatory documents[18]. Nevertheless, LLM-only pipelines remain limited by hallucination, inconsistent recall, and insufficient grounding in authoritative biomedical knowledge, which limit their reliability for large-scale dataset construction.

Retrieval-Augmented Generation (RAG) has emerged as an effective paradigm for addressing these limitations by explicitly conditioning LLM outputs on retrieved external evidence[19]. In the biomedical domain, RAG-based systems such as BioMedRAG[20], MedicalRAG[21], and MedRAG[22] integrate relevant literature to improve factual accuracy and downstream task performance. Knowledge Graph (KG)-enhanced RAG approaches, including DALK[23] and KRAGEN[24], further strengthen retrieval and reasoning by organizing biomedical entities and relationships into structured representations that support contextual grounding and relational reasoning. More general frameworks, such as LightRAG[25], MiniRAG[26], GraphRAG[27], and ArchRAG[28], demonstrate scalable mechanisms for integrating retrieval, structured knowledge, and LLM-based reasoning across domains. Despite these advances, existing RAG-based methods have not been systematically designed to construct comprehensive, multi-task Drug-Disease datasets that simultaneously support DDM, DEA, and DTA. This gap limits the availability of unified, reusable resources for downstream AI-driven biomedical research.

To address this challenge, we introduce $UniD^3$, a unified framework that integrates Large Language Models with Knowledge Graph-enhanced Retrieval-Augmented Generation (KG-RAG) to systematically extract drug-disease relationships from biomedical literature, organize them into structured knowledge graphs, and expose the resulting knowledge through curated datasets and an interactive question-answering interface. $UniD^3$ employs a dual-stage entity extraction and validation strategy that progressively contructs and refines biomedical knowledge graphs, enabling accurate, scalable, and robust drug-disease knowledge organization. Using $UniD^3$, we develop a comprehensive $D^3$ resource spanning DDM, DEA, and DTA tasks, validated by clinicians and cross-referenced with external curated datasets. In addition, we benchmark multiple general-purpose and biomedical-specific LLMs on the $UniD^3$-generated dataset to assess their performance in Drug-Disease question answering and related tasks. Collectively, $UniD^3$ provides a scalable and extensible foundation for transforming unstructured biomedical literature into high-quality, structured drug-disease knowledge. To make this knowledge broadly accessible, we further develop an explainable web-based chatbot (https://unid3.site/), which allows researchers and clinicians to interactively explore drug-disease associations with evidence-grounded responses and traceable citations. Together, the knowledge graphs, datasets, and chatbot form an integrated platform for AI-driven drug discovery, drug repurposing, and precision medicine. All resources are publicly available through the $UniD^3$ project (https://github.com/QSong-github/UniD3).

# METHODS

## Overview of $UniD^3$ framework

We present $UniD^3$, a unified framework that integrates Large Language Models (LLMs) with Knowledge Graph-enhanced Retrieval-Augmented Generation (KG-RAG) to systematically extract and organize drug-disease relationships from biomedical literature. $UniD^3$ bridges the gap between unstructured biomedical text and structured, machine-readable Drug-Disease knowledge through a dual-stage entity extraction strategy that enhances extraction accuracy, scalability, and cross-document consistency.

As shown in **Figure 1a**, UniD$^3$ begins with large-scale collection of PubMed articles related to drug-centered biomedical research. Retrieved articles are processed using the Llama 3.3-70B model and classified via a task-specific research classification prompt into one of three core tasks: Drug-Disease Matching (DDM), Drug Effectiveness Assessment (DEA), or Drug-Target Analysis (DTA), ensuring task relevance and minimizing noise from unrelated studies. After task assignment, UniD$^3$ performs entity and relationship extraction focused on key biomedical concepts, including drugs, diseases, molecular targets, pathways, and clinical outcomes. To enhance robustness and contextual completeness, UniD$^3$ adopts a dual-stage extraction strategy. In Stage I (paper-level extraction), each article is first segmented into manageable textual units. Within each segment, UniD$^3$ independently identifies explicit entity mentions such as drug names, disease terms, protein targets, and pathway identifiers, and extracts directly stated task-specific relationships, including drug-disease associations, drug–target interactions, and target-pathway links. These extracted items represent text-local biomedical facts grounded strictly in the source passages without graph-level aggregation. In Stage II (knowledge graph integration), extracted entities are aggregated through one-hop neighborhood expansion and knowledge graph consolidation. Drug and Disease entities are retained as anchor nodes, and each anchor is expanded by incorporating its one-hop neighbors along with curated textual descriptors and keyword annotations. This process produces semantically enriched entity representations that integrate corroborating evidence across documents and capture higher-order biomedical relationships beyond individual text contexts. Task-specific dataset generation prompts (**Figure 2**) are then applied to generate structured datasets for DDM, DEA, and DTA using Llama 3.3-70B. The resulting datasets are evaluated through automated quality checks, cross-referencing with external curated resources, and expert clinical review. These curated datasets further support downstream benchmarking of general-purpose and biomedical-specific LLMs in Drug-Disease question answering tasks, enabling systematic assessment of model performance and practical applicability.

The resulting knowledge graph and structured datasets enable the explainable biomedical chatbot interface (https://unid3.site/) illustrated in **Figure 1b**. Through KG-RAG retrieval, this conversational agent answers drug-disease queries by integrating knowledge graph context with supporting literature evidence. The chatbot provides interpretable, evidence-grounded responses with traceable citations and supports interactive exploration of drug indications, therapeutic effectiveness, and molecular mechanisms via natural language dialogue.

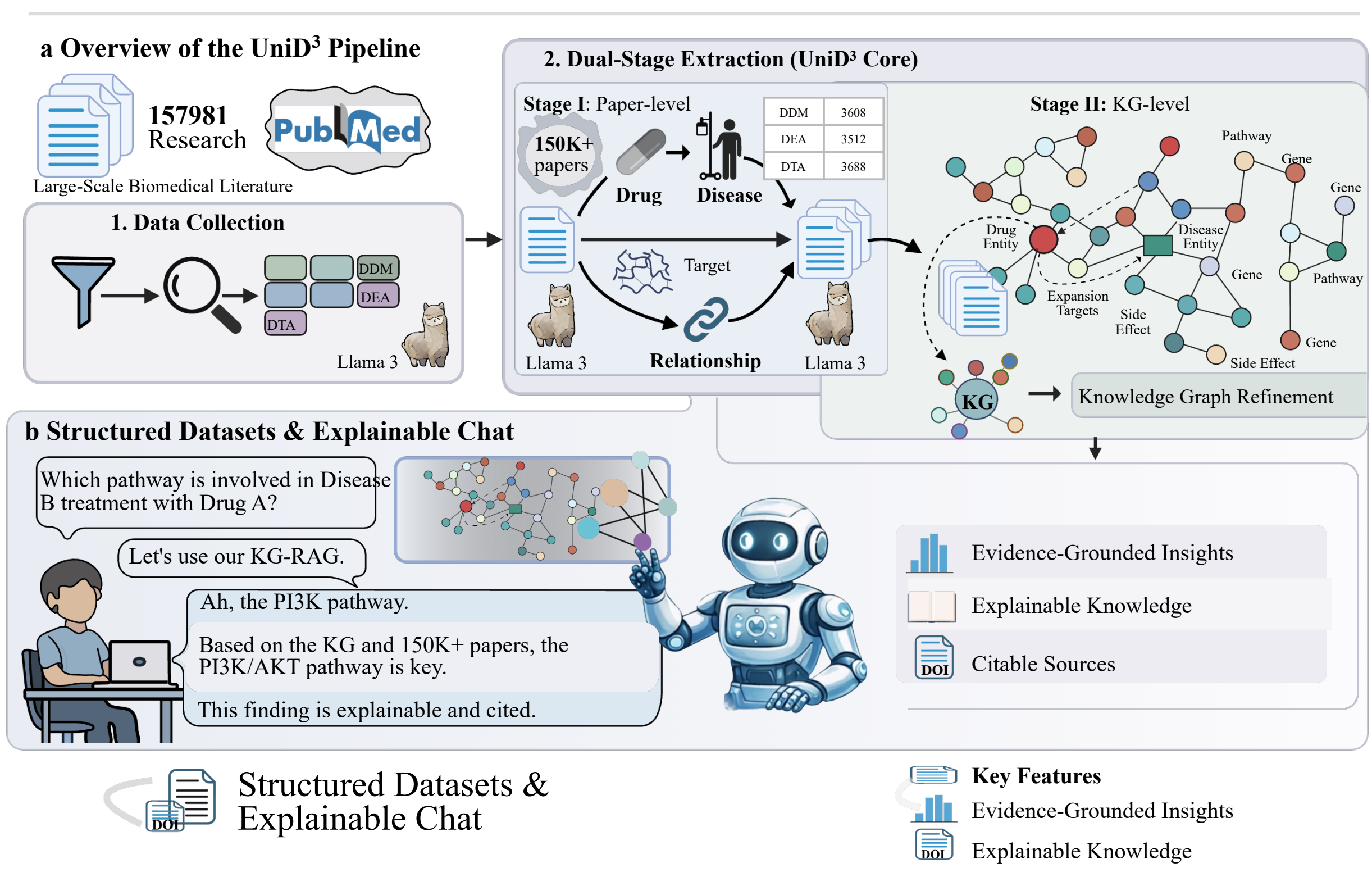

**Figure 1**: **Overview of the UniD³ framework for KG construction, dataset generation, and explainable question answering**. **(a)** Overview of the UniD³ framework. PubMed articles are collected and classified by Llama 3.3 into three tasks: Drug–Disease Matching (DDM), Drug Effectiveness Assessment (DEA), and Drug-Target Analysis (DTA). UniD³ then performs a dual-stage extraction process. In Stage I (paper-level), task-specific prompts identify biomedical entities and extract relational evidence from individual articles. In Stage II (knowledge graph/KG-level), the extracted entities and relations are consolidated into a biomedical knowledge graph and used by LightRAG to generate datasets for the DDM, DEA, and DTA tasks. **(b)** Structured datasets and explainable chatbot. The UniD³ generated KG and datasets support an evidence-grounded conversational interface for biomedical queries.

## Research Article Retrieval and Preprocessing

We selected research articles from PubMed (https://pubmed.ncbi.nlm.nih.gov/?term=drug) using the keyword 'Drug' and downloaded all available publications up to September 2024, totaling 157,849 papers. Task-oriented prompts were then used to guide the Llama3.3-70B model in identifying relevant studies. Specifically, we designed three task-specific prompts corresponding to Drug-Disease Matching, Drug Effectiveness Assessment, and Drug-Target Analysis (**Figure 2**) and applied them to perform task-specific classification and filtering. Using his approach, the Llama3.3-70B model initially identified 4,701, 4,642, and 4,638 papers for the three tasks, respectively. However, due to potential hallucination effects in the LLM, some of the extracted titles were incomplete or inconsistent with the expected content. To address this, we applied fuzzy string matching[29] to restore incomplete titles and remove unmatched entries. Titles with a similarity score below 0.5 were excluded. After refinement, we obtained 3,608, 3,512, and 3,688 papers for the three tasks, respectively.

**Research classification prompt**

You should think as an expert in the medical field, I will give you a bunch of research paper titles. Please select the appropriate papers for the three tasks I set. You need to classify each paper title. Of course, if you think a paper does not belong to any task, you can classify it as "irrelevant". The following are the specific descriptions of the three tasks:

Task1: Drug Disease Matching

Identify potential drugs that could treat [specific disease]. For each drug, provide an analysis of its target genes and pathways and explain their relevance to the disease mechanism.

Task2: Drug Effectiveness Assessment

Evaluate the effectiveness of [specific drug] for treating [specific disease]. Include an analysis of clinical or preclinical data, the strength of the drug's interaction with its target genes or pathways, and any evidence of therapeutic outcomes.

Task3: Drug Target Analysis

Map the genes and pathways targeted by [specific drug]. Explain how these targets are implicated in [specific disease] and assess whether the drug's mechanism of action aligns with the disease's underlying biology.

You must give the result in the following format:

Drug Disease Matching:[title1,title2,title3,title4,...]

Drug Effectiveness Assessment:[title1,title2,title4,title4,...]

Drug Target Analysis:[title1,title2,title3,title4,...]

Now think carefully and categorize these titles carefully.

The following are the titles that need to be categorized:

**Figure 2: Research classification prompt used in UniD³.** Example of the task-specific prompt provided to Llama 3.3-70B for classifying biomedical research articles into Drug-Disease Matching (DDM), Drug Effectiveness Assessment (DEA), and Drug-Target Analysis (DTA). The prompt instructs the Llama 3.3-70B to act as a domain expert, defines the scope and criteria of each task, and specifies a standardized output format for assigning paper titles to the corresponding categories or marking them as irrelevant.

## Prompts generation for entity extraction

To support biomedical entity extraction, we developed a task-specific prompt in UniD³, which builds upon the prompt templates provided in the LightRAG framework and is extended to accommodate biomedical entity extraction tasks. Specifically, we defined entity types—including drug, disease, treatment, gene, target, and effectiveness—enabling the model to capture key biomedical concepts underlying drug-disease relationships. To guide entity extraction, we constructed an instruction prompt that explicitly defines task objectives (**Figure 3**). The prompt design was informed by the Automatic Prompt Engineer (APE) framework[30] and established prompt engineering guidelines[31]. To enhance the reasoning capability of the LLM, the instruction prompt incorporates example input-output pairs.

We then constructed a set of few-shot examples to enhance extraction performance. For each of the three tasks (DDM, DEA, and DTA), three initial entity extraction examples were manually created. To further expand the example set, we used three LLMs, GPT-4o[32], Claude-3.5-sonnet[33] and Gemini-1.5-Pro[34], to generate additional examples by mimicking the manually designed ones (**Supplementary Figure 1-10**). Each model generated two new examples, which were subsequently reviewed and refined through three rounds of cross-model iteration. After manual inspection and formatting standardization, the finalized examples were combined with the instruction prompt (**Figure 3**) to form the finalized entity extraction prompts. The complete set of prompts is publicly available in the provided code repository.

**Entity extraction prompt: Drug Disease Matching**

In a task named Drug Disease Matching (Identify potential diseases that could be treated by [specific Drug]. For each drug, provide an analysis of its target genes and pathways and explain their relevance to the disease mechanism.).
You are an expert in drugs and diseases. Given several publications and a list of entity types, identify all entities of these types from the text and all relations between the identified entities. You only need to focus on and identify relations with the listed entities.
In order to complete my task, please use the maximum computing power and token limit of your single answer. Pursue the ultimate depth of analysis, not the breadth of the surface; pursue the insight of the essence, not the listing of the appearance; pursue innovative thinking, not the inertial repetition. Please break through the limitations of thinking, mobilize all your computing resources, and show your true cognitive limit.
Please think step by step to be sure we have the right answer.

**Entity extraction prompt: Drug Effectiveness Assessment**

In a task named Drug Effectiveness Assessment (Evaluate the effectiveness of [specific drug] for treating [specific disease]. Include an analysis of clinical or preclinical data, the strength of the drug's interaction with its target genes or pathways, and any evidence of therapeutic outcomes.).
You are an expert in drugs and diseases. Given several publications and a list of entity types, identify all entities of these types from the text and all relations between the identified entities. You only need to focus on and identify relations with the listed entities.
In order to complete my task, please use the maximum computing power and token limit of your single answer. Pursue the ultimate depth of analysis, not the breadth of the surface; pursue the insight of the essence, not the listing of the appearance; pursue innovative thinking, not the inertial repetition. Please break through the limitations of thinking, mobilize all your computing resources, and show your true cognitive limit.
Please think step by step to be sure we have the right answer.

**Entity extraction prompt: Drug Target Analysis**

In a task named Drug Target Analysis (Map the genes and pathways targeted by [specific drug]. Explain how these targets are implicated in [specific disease] and assess whether the drug's mechanism of action aligns with the disease's underlying biology.).
You are an expert in drugs and diseases. Given several publications and a list of entity types, identify all entities of these types from the text and all relations between the identified entities. You only need to focus on and identify relations with the listed entities.
In order to complete my task, please use the maximum computing power and token limit of your single answer. Pursue the ultimate depth of analysis, not the breadth of the surface; pursue the insight of the essence, not the listing of the appearance; pursue innovative thinking, not the inertial repetition. Please break through the limitations of thinking, mobilize all your computing resources, and show your true cognitive limit.
Please think step by step to be sure we have the right answer.

**Figure 3: Task-specific entity extraction prompts used in UniD$^3$.** Illustration of the prompt templates provided to Llama 3.3-70B for entity and relationship extraction in the three UniD$^3$ tasks: Drug-Disease Matching (DDM), Drug Effectiveness Assessment (DEA), and Drug-Target Analysis (DTA). Each prompt defines the task objective, specifies relevant entity types, and instructs Llama 3.3-70B to extract task-specific entities and relationships from biomedical publications, enabling consistent and structured knowledge extraction across tasks.

### Dual-stage entity extraction strategy

Entity extraction for knowledge graph (KG) construction is a critical step in UniD$^3$. By leveraging the LLM and LightRAG framework, we efficiently extract relevant information and organize it into structured triples. Despite the robust entity extraction capabilities of both LightRAG and the LLM, we implemented a dual-stage entity extraction strategy to address computational cost and information dispersion. From a computational perspective, the number of extracted entities substantially exceeds the number of source articles. Fine-grained segmentation into small semantic units would generate a large number of text chunks, significantly increasing processing costs. To mitigate this, we adopt a larger chunking strategy to to reduce computational overhead. Accordingly, in Stage I (Paper-Level Entity Extraction) of entity extraction, we set the chunk size to 8,000 tokens. While larger chunks improve processing efficiency, they may increase the risk of missing localized information. However, given that valuable information in biomedical literature is relatively dispersed and irrelevant content constitutes the majority, this chunk size effectively balances processing efficiency and information retention.

Due to the dispersion of information, the LLM lacks access to cross-chunk and cross-document when processing individual text segments. However, the dual-stage entity extraction strategy addresses this limitation by aggregating entity information in Stage II (KG-Level Entity Extraction), providing the LLM with structred and contextually enriched knowledge. In the second stage, we refine and consolidate the triples extracted in stage I. Specifically, we retain only the Drug and Disease entities as anchors and augment each entity with its one-hop neighboring nodes, along with relevant descriptions and keywords, to construct entity-centric summaries. Each summary is treated as a distinct chunk, and processed through the LightRAG framework in stage II, enhanced entity extraction and knowledge graph construction with improved contextual completeness.

### UniD$^3$ Dataset generation

For the dataset generation, we use Llama 3.3-70B as the generator, integrated within the LightRAG framework. For all three tasks, we adpot the "Mix" retrieval strategy, which combines knowledge graph and vector-based retrieval to provide context-rich and comprehensive inputs. This approach leverages both structured and unstructured information, enabling the model to interpret drug-related knowledge more effectively and produce high-quality outputs. For the Drug-Disease Matching task, Llama 3.3-70B is provided with a drug name and instructed to generate a comprehensive set of associated diseases, focusing on those with potential therapeutic relevance. The DDM dataset is formulated as an open multi-label classification problem, with applications in drug recommendation repurposing. The Drug Effectiveness Assessment (DEA) task is structured as a binary classification problem, where the model generates disease list and labels each as effective or ineffective, accompanied by supporting explanations. The DEA dataset supports automated effectiveness screening, evidence synthesis, and clinical decision-making support. The Drug-Target Analysis (DTA) task is designed as an open-ended question-answering dataset, where Llama 3.3-70B generate multiple drug-related question-answer pairs given drug name. The format of the QA pairs is unconstrained, allowing for more flexible knowledge elicitation. These UniD$^3$ generated datasets can serve as a valuable resource (i.e. biomedical QA) for biomedical information retrieval, knowledge base construction, and drug education interfaces. The dataset generation prompts and detailed instructions for each task are provided in **Supplementary Figure 11**.

## RESULTS

### UniD$^3$ Constructed Knowledge Graph

Using the UniD$^3$ framework with Llama 3.3-70B, we constructed six knowledge graphs spanning two extraction stages (Stage I: paper-level, Stage II: KG-level) and three biomedical tasks (DDM, DEA, DTA). **Table 1** summarizes the statistics of the extraction pipeline, and **Figure 4** presents the structural characteristics of the resulting KGs.

The extraction pipeline begins with raw research documents, i.e. 3,608 (DDM), 3,512 (DEA), and 3,688 (DTA) papers, which are first segmented into text chunks (5,800, 4,700, and 5,800, respectively), totaling 22.1M–30.4M tokens. In Stage I, paper-level extraction yields large but sparse KGs: DDM-KG contains 64,300 nodes and 72,000 edges, DEA-KG contains 56,900 nodes and 63,400 edges, and DTA-KG contains 69,900 nodes and 74,000 edges (**Table 1**). Stage II applies refined re-extraction on a smaller set of more information-dense chunks (13,000–16,700 chunks with 2.2M–4.0M tokens), producing consolidated graphs with fewer nodes (30,500–45,900) but considerably more edges (84,100–126,000). Critically, the KG density increases by approximately one order of magnitude from Stage I to Stage II (e.g., from $3.48 \times 10^{-5}$ to $1.20 \times 10^{-4}$ for DDM), confirming that the re-extraction process effectively distills sparser raw graphs into denser, higher-quality knowledge representations.

**Figure 4a** presents the entity type composition across all six KGs. In Stage I, the "Other" category (encompassing biological processes, pathways, and biomarkers) dominates across all tasks (63.3% - 74.1%), reflecting the breadth of concepts captured during paper-level extraction. After Stage II consolidation, the proportion of task-relevant entities increases substantially: Drug entities rise from 14.1% to 21.3% in DDM-KG, Disease entities increase from 8.6% to 12.7% in DEA-KG, and Gene/Target entities grow from 15.6% to 23.7% in DTA-KG, while the "Other" fraction decrease to 44.6% - 47.8% across all three tasks. These shifts demonstrate that Stage II consolidation effectively enriches domain-specific knowledge while filtering peripheral concepts. **Figure 4b** depicts the degree distribution of all six KGs on log-log axes. Both stages exhibit power-law-like distributions, a hallmark of scale-free networks commonly observed in biological knowledge graphs. Stage I KGs are sparse, with average degrees of 2.2 (DDM), 2.2 (DEA), and 2.1 (DTA), whereas Stage II KGs achieve substantially higher connectivity, with average degrees of 5.5 (DDM), 5.5 (DEA), and 5.2 (DTA), with heavier distribution tails reflecting the increased high-connectivity hubs formed through KG-level consolidation. **Figure 4c** further quantifies the structural transition across three complementary metrics. Average node degree increases by 2.4 – 2.5-fold from Stage I to Stage II across all tasks, confirming a consistent densification effect. The largest connected component ratio expands from approximately 59 – 61% in Stage I to over 92% in Stage II across all three KGs, indicating that the majority of entities become integrated into a single navigable network. Correspondingly, the total number of connected components decreases sharply by 6.9-fold in DDM (from 19,600 to 2,800), 8.7-fold in DEA (from 16,000 to 1,800), and 9.2-fold in DTA (from 21,500 to 2,300), confirming that KG-level re-extraction successfully bridges fragmented knowledge clusters across papers into a cohesive and navigable graph structure.

**Table 1 Statistics of the dual-stage entity extraction pipeline across the DDM, DEA, and DTA knowledge graph.** Stage I performs initial chunking and extraction from raw research documents, while Stage II applies refined re-extraction to yield consolidated knowledge graph components. For each stage, we report the number of text chunks, total tokens, extracted nodes, edges, and the resulting KG density. Notably, Stage II produces substantially fewer but more concentrated chunks with increased KG density (approximately one order of magnitude higher), indicating that the re-extraction step effectively transfroms sparser raw graphs into denser, higher-quality knowledge representations.

| Dataset | Research | Stage I | | | | | Stage II | | | | |
|---|---|---|---|---|---|---|---|---|---|---|---|
| | | Chunk | Token | Node | Edge | KG density | Chunk | Token | Node | Edge | KG density |
| **DDM** | 3.6K | 5.8K | 30.4M | 64.3K | 72.0K | 3.48E-05 | 16.7K | 3.4M | 45.9K | 126.0K | 1.20E-04 |
| **DEA** | 3.5K | 4.7K | 22.1M | 56.9K | 63.4K | 3.91E-05 | 108K | 2.3M | 30.5K | 84.1K | 1.81E-04 |
| **DTA** | 3.7K | 5.8K | 29.6M | 69.9K | 74.0K | 3.03E-05 | 13.0K | 2.2M | 36.3K | 93.5K | 1.42E-04 |

## Drug-Centric Knowledge Exploration

To illustrate the practical utility of UniD$^3$ constructed KGs, we present a drug-centric knowledge exploration centered on Cisplatin, a widely used platinum-based chemotherapy agent[35], within the Stage II DDM-KG (**Figure 5**). **Figure 5a** visualizes the ego-network of Cisplatin, displaying entities within two hops of the central Drug node. The subgraph reveals a richly connected knowledge landscape: Cisplatin is directly linked to a broad spectrum of Disease entities including Lung Cancer, Breast Cancer, Colorectal Cancer, Glioblastoma, Ovarian Cancer, and Melanoma, reflecting its extensive clinical applications across oncology. Beyond direct drug-disease associations, the graph captures Gene/Target entities such as EGFR and treatment-related concepts including Cancer Treatment, Chemotherapy, Drug Resistance, and Apoptosis, forming a multi-layered representation of Cisplatin's pharmacological context[36]. The presence of resistance-related entities (Drug Resistance, Resistance, Multidrug Resistance) alongside mechanistic nodes (DNA Damage, DNA Replication, Tumor Suppression) demonstrates that UniD$^3$'s KGs encode not only therapeutic associations but also underlying biological mechanisms and clinical challenges.

**Figure 5b** extends this exploration by presenting curated multi-hop relationship paths for four representative drugs (i.e. Imatinib, Trastuzumab, Sorafenib, and Metformin), each illustrating mechanistically distinct route types encoded within the KG. For Imatinib, the two-hop path linking it to Chronic Myeloid Leukemia (CML) through the BCR-ABL Gene captures its canonical mechanism as a BCR-ABL tyrosine kinase inhibitor[37], while a parallel path to Ph+ ALL through the same intermediate reflects the shared oncogenic etiology across Philadelphia chromosome-positive leukemias; a third path routed through Cell Cycle Arrest and P53 to Breast Cancer reveals a secondary, target-agnostic axis supporting exploratory use beyond hematologic indications. For Trastuzumab, the direct two-hop connection to Breast Cancer via

HER2 encodes its primary approved indication, while paths mediated by the Monoclonal Antibody class node (leading to HER2-driven Breast Cancer and to EGFR-driven NSCLC), contextualize its mechanism within the broader class-effect landscape of anti-ErbB therapies[38]. For Sorafenib, a two-hop path linking it to Melanoma through the BRAF Gene reflects its multi-kinase inhibitory activity against the BRAF-driven MAPK cascade[39], whereas two ferroptosis-mediated three-hop paths, connecting it to Gastric Cancer via FTH1 (Ferritin Heavy Chain 1)[40], and to Breast Cancer via P53, delineate an orthogonal oxidative cell death mechanism. For Metformin, the KG encodes both its primary indication through a direct edge to Type 2 Diabetes and two repurposing-relevant axes: a host-directed antimicrobial path reaching Tuberculosis via Mycobacterium tuberculosis, and an oncologic path connecting it to Prostate Cancer through Inhibition and the Androgen Receptor[41]. Collectively, these structured pathways demonstrate that UniD$^3$ KGs encode not only direct drug-disease associations but also multi-hop mechanistic and translational relationships that underpin hypothesis generation for drug repurposing and mechanism-of-action studies.

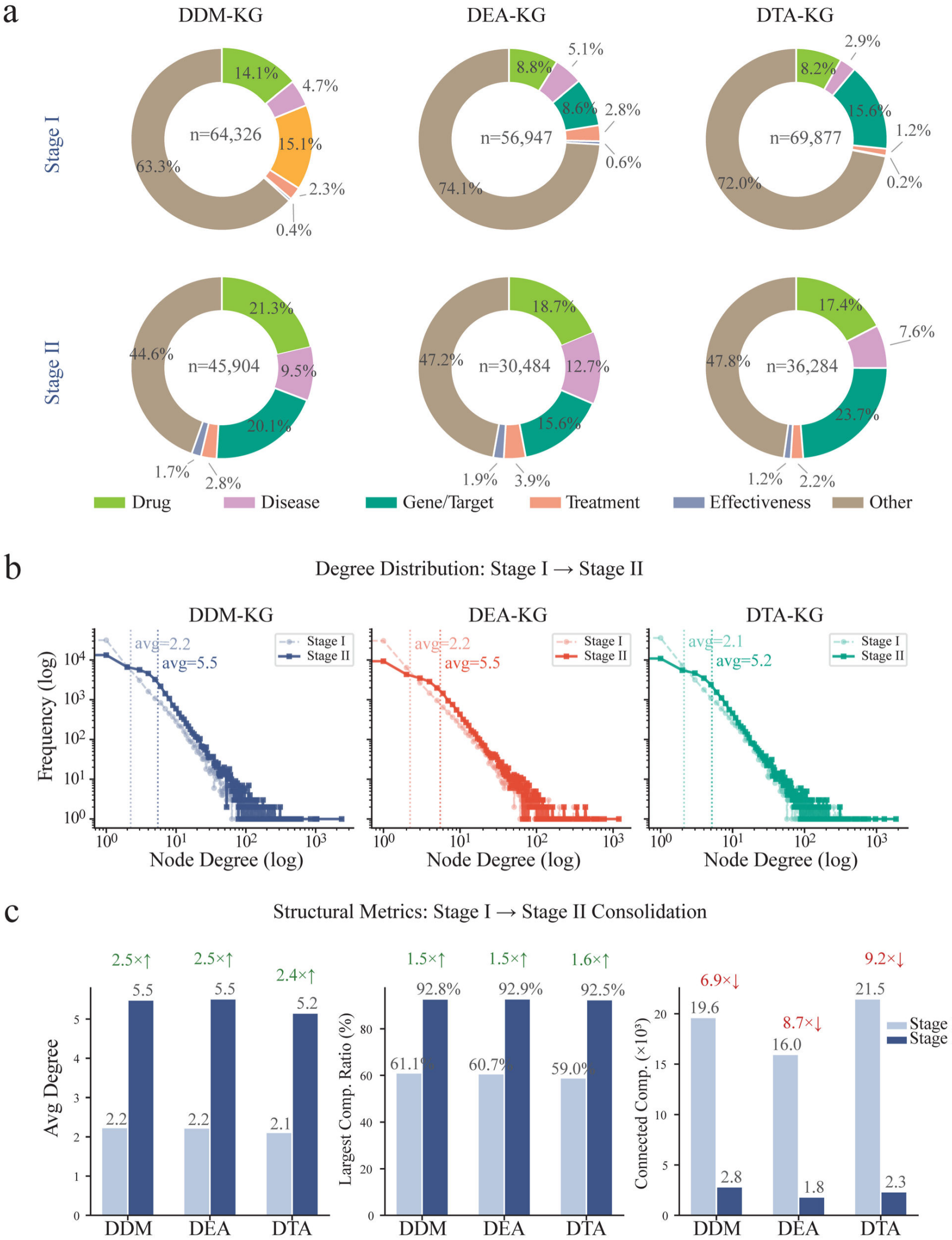

**Figure 4 Structural overview of the UniD$^3$ knowledge graphs. (a)** Entity type composition across the six KGs (two stages for three tasks). Each chart displays the proportion of Drug, Disease, Gene/Target, Treatment, Effectiveness, and Other entities; the total node count is annotated at the center. Stage II graphs exhibit substantially enriched proportions of task-relevant entity types and reduced "Other" fractions relative to Stage I. **(b)** Degree distributions of the Stage I and Stage II KGs on log-log axes for DDM, DEA, and DTA. All graphs follow power-law-like distributions characteristic of scale-free networks; Stage II KGs display higher average degrees (5.2 - 5.5 vs. 2.1 - 2.2) and heavier tails, reflecting increased connectivity through KG-level consolidation. **(c)** Structural metrics comparing Stage I and Stage II across three indicators: average node degree (2.4-2.5x increase), largest connected component ratio (expanding to > 92%), and number of connected components (6.9-9.2x decrease). These metrics confirm that Stage

II consolidation transforms sparse, fragmented paper-level graphs into dense, globally connected knowledge structures.

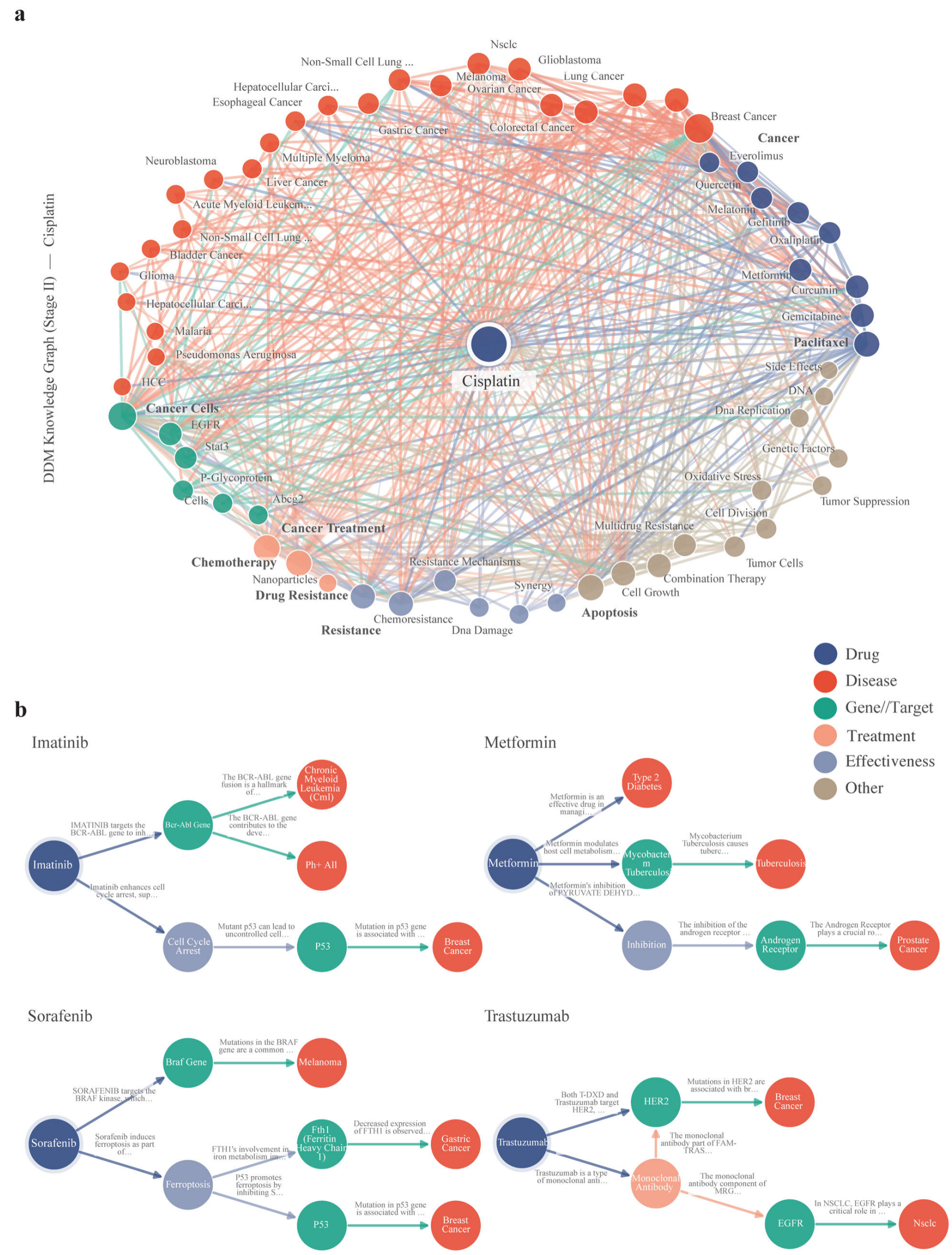

**Figure 5 Drug-centric knowledge exploration in the DDM knowledge graph. (a)** Ego-network of Cisplatin in the Stage II DDM-KG, visualizing entities within two hops of the central node. Nodes are colored by entity type (Drug, Disease, Gene/Target, Treatment, Effectiveness, Other) and scaled by degree. **(b)** Curated relationship paths for four representative drugs (Imatinib, Trastuzumab, Sorafenib, and Metformin) in the Stage II DDM-KG, illustrating mechanistically distinct multi-hop route types. Each path connects a drug through one or more intermediate biological entities (e.g. BCR-ABL Gene, Monoclonal Antibody, Ferroptosis, or Inhibition) to a downstream disease endpoint. Edge annotations indicate extracted relationship descriptions. Paths were selected to reflect canonical indications, shared mechanism class effects, and drug repurposing candidates.

## UniD$^3$ Generated Dataset

The UniD$^3$-generated datasets include three task-specific datasets: Drug-Disease Matching (DDM), Drug Effectiveness Assessment (DEA), and Drug-Target Analysis (DTA). The DDM dataset is an open multi-label classification task comprising 28,915 samples encompassing 8,576 drugs. The DEA dataset is structured as a binary classification task, consisting of 15,042 samples involving over 3,000 drugs. The DTA dataset is formulated as a general question-answering (QA) task, containing more than 4,000 drug-related question-answer pairs.

For the validation of the UniD$^3$-generated dataset, we used multiple external datasets from Hugging Face[42], including CDR[43], DrugReviews[44], P3ps[45] and MedicationQA[46]. Among these, CDR, DrugReviews, and P3ps are utilized to validate DDM and DEA, while MedicationQA is employed to validate DTA. The external dataset validation results are shown in **Table 2**. In the DDM dataset, given the sparsity of exact matches between drug-disease pairs, we employed fuzzy matching techniques to align drug-disease pairs, identifying a total of 831 matched pairs across the three datasets. On the CDR dataset, DDM achieved a precision of 0.8302, recall of 0.8778, and F1 score of 0.8526. Using DrugReviews, ratings of 7 or higher were considered effective, and ratings of 3 or lower were considered ineffective. For DrugReviews, we obtained a precision of 0.8553, recall of 0.8923, and F1 score of 0.8727, while on the P3ps dataset, the corresponding scores were 0.8486, 0.8951, and 0.8707. In DEA, we followed a similar fuzzy matching strategy and identified 855 drug-disease pairs across the datasets. The DEA dataset exhibited consistentlystrong performance across all datasets, with precision, recall, and F1 scores of 0.8495, 0.8851, and 0.8658 on CDR, 0.8553, 0.8923, and 0.8727 on DrugReviews_effective, 0.8405, 0.8846, and 0.8615 on DrugReviews_ineffective, and 0.8494, 0.892, and 0.8696 on P3ps. For DTA evaluation using the MedicationQA dataset, 190 matched drug-target-related Q&A pairs achieved a precision of 0.8147, recall of 0.8287, and F1 score of 0.8214.

To further verify the quality of the UniD$^3$-generated DDM dataset, we invited clinicians to review the dataset after filtering out drugs that are FDA-approved and currently marketed in the United States, leaving 469 unique drugs. Despite minor spelling errors, most drug names could be matched to official names. Clinical experts assessed whether each drug was labeled as "effective" or "ineffective" based on available evidence. The verification results showed an accuracy of 0.9000, precision of 0.8810, recall of 0.9209, F1 score of 0.9005, and AUROC of 0.9004, indicating strong agreement between expert assessments and dataset annotations, thereby demonstrating the dataset's robustness and suitability for drug effectiveness classification tasks.

**Table 2 External validation performance of DDM, DEA, and DTA across four benchmark datasets including CDR, DrugReviews, P3ps, and MedicationQA, evaluated using Precision, Recall, and F1 score.** DDM and DEA are assessed on retrieval-oriented tasks covering chemical–disease relations (CDR), patient drug reviews (DrugReviews), and biomedical literature passages (P3ps), while DTA is exclusively evaluated on MedicationQA for open-ended drug question answering.

| | DDM | | | DEA | | | DTA | | |
|---|---|---|---|---|---|---|---|---|---|
| | Precision | Recall | F1 | Precision | Recall | F1 | Precision | Recall | F1 |
| **CDR** | 0.8302 | 0.8778 | 0.8526 | 0.8495 | 0.8851 | 0.8668 | - | - | - |
| **DrugReviews** | 0.8553 | 0.8923 | 0.8727 | 0.8479 | 0.8885 | 0.8677 | - | - | - |
| **P3ps** | 0.8486 | 0.8951 | 0.8707 | 0.8494 | 0.892 | 0.8696 | - | - | - |
| **MedicationQA** | - | - | - | - | - | - | **0.8147** | **0.8287** | **0.8214** |

## Benchmark of UniD$^3$-generated datasets

We benchmarked the UniD$^3$-generated datasets by evaluating five biomedical large language models including Llama3-OpenBioLLM (70B)[47], MedLlama3 (8B)[48], MedX (8B)[49], Huatuo (7B)[50], BioGPT (1.6B)[51] and two general large language models including GPTo4-mini[32] and Llama3.3 (70B)[52]. In addition, we tested whether the system can reliably reproduce its own results by querying UniD$^3$ itself. Some of the five biomedical models rank among the top performers on the Open Medical LLM Leaderboard[53].

The results of the benchmark are shown in **Table 3**. Unsurprisingly, Llama3.3-LightRAG-mix (based on the Llama3.3 model, using the mix retrieval method under the LightRAG framework) achieves the best overall performance, demonstrating the self-consistency and effectiveness of retrieval-augmented generation. Beyond the LightRAG-augmented models, the open multi-label classification task (DDM) reveals notable performance variation among standalone models. For each drug, models generates a list of associated diseases, which are evaluated against ground-truth annotatons. Cov1 measures how often at least one prediction is correct in the whole list, Cov2 requires at least two correct predictions, and Cov3 requires at least three. Among all non-RAG baselines, Llama3.3 (70B) achieves the strongest DDM performance, reaching 0.7981, 0.6298, and 0.4311 on Cov1, Cov2, and Cov3 respectively, suggesting that larger general-purpose models can partially compensate for the absence of retrieval augmentation through broader parametric knowledge. In the binary classification task (DEA), performance is evaluated based on drug effectiveness prediction accuracy. Llama3-OpenBioLLM (70B) has outstanding performance in accuracy (0.8277), F1 (0.8305), and AUC (0.8657). In the open question answering task (DTA), evaluation focus on text similarity and generation quality matrics. Llama3.3 demonstrates strong performance across KWM, BLEU, ROUGE and BERTsc, especially reaching 0.9092 on BERTsc.

**Table 3 Comprehensive benchmarking of seven baseline large language models and three LightRAG-augmented variants on DDM, DEA, and DTA tasks.** We evaluate seven standanlone language models and three UniD$^3$ querying modes based on Llama3.3-70B: local (context-dependent retrieval), global (global knowledge retrieval) and mix (integration of knowledge graph and vector retrieval; the mode used for dataset generation). Cov1, Cov2, and Cov3 denote the proportion of samples for which at least 1, 2, and 3 predicted diseases match the ground-truth set, respectively. As the minimum required number of matched diseases increases, the metric becomes progressively more stringent, yielding monotonically decreasing scores from Cov1 to Cov3. SS denotes the Semantic Similarity, KWM is the Keyword Matching Score[54], BLEU is the BLEU Score[55], ROUGE refers to the ROUGE-L Score[56], and BERTsc is the BERTScore[57].

| Dataset | | DDM | | | DEA | | | | DTA | | |
|---|---|---|---|---|---|---|---|---|---|---|---|
| **Metrics** | **Cov1** | **Cov2** | **Cov3** | **ACC** | **F1** | **AOC** | **SS** | **KWM** | **BLEU** | **ROUGE** | **BERTsc** |
| **Llama3-OpenBioLLM (70B)** | 0.7861 | 0.7015 | 0.5133 | 0.8277 | 0.8305 | 0.8657 | 0.8428 | 0.3811 | 0.4343 | 0.3977 | 0.8591 |
| **BioGPT (1.6B)** | 0.4211 | 0.2879 | 0.1613 | 0.4964 | 0.6072 | 0.5311 | 0.5183 | 0.2275 | 0.2345 | 0.2079 | 0.8241 |
| **Huatuo (7B)** | 0.4577 | 0.3195 | 0.2006 | 0.5598 | 0.6726 | 0.5883 | 0.2838 | 0.2407 | 0.241 | 0.2361 | 0.8273 |
| **MedX (8B)** | 0.6827 | 0.4551 | 0.3635 | 0.6544 | 0.5963 | 0.6514 | 0.3667 | 0.2073 | 0.2337 | 0.2894 | 0.8169 |
| **MedLlama3 (8B)** | 0.5177 | 0.3872 | 0.2119 | 0.5753 | 0.6058 | 0.5778 | 0.3101 | 0.1818 | 0.2461 | 0.2511 | 0.8378 |
| **GPTo4-mini (NA)** | 0.7522 | 0.6332 | 0.5001 | 0.793 | 0.7568 | 0.7895 | 0.8112 | 0.3962 | 0.4223 | 0.449 | 0.8682 |
| **Llama3.3 (70B)** | 0.7981 | 0.6298 | 0.4311 | 0.7641 | 0.7762 | 0.7662 | 0.8834 | 0.4695 | 0.5435 | 0.4315 | 0.9092 |
| **Llama3.3-LightRAG-local** | 0.8437 | 0.7958 | 0.7655 | 0.8491 | 0.8627 | 0.8433 | 0.8957 | 0.6715 | 0.5366 | 0.6385 | 0.9418 |
| **Llama3.3-LightRAG-global** | 0.8967 | 0.8325 | 0.7968 | 0.8635 | 0.8589 | 0.8372 | 0.9369 | 0.6997 | 0.6218 | 0.6739 | 0.9677 |
| **Llama3.3-LightRAG-mix** | 0.9437 | 0.8705 | 0.8388 | 0.9764 | 0.9883 | 0.9899 | 0.9811 | 0.8497 | 0.7815 | 0.7568 | 0.9871 |

## UniD$^3$ Chatbot and Query Case Studies

To facilitate intuitive interaction with extracted biomedical knowledge, we developed the UniD$^3$ Chatbot, a web-based question-answering system available at https://unid3.site/. The system integrates large language models (LLMs) with structured biomedical knowledge graphs constructed through literature mining and entity–relationship extraction. Users can query complex biomedical associations (e.g. drug–disease relationships, gene interactions, treatment mechanisms, and biological insights) using natural language. Unlike conventional retrieval-only systems, UniD$^3$ Chatbot combines structured retrieval with generative reasoning. Specifically, it first retrieves relevant facts and relationships from the knowledge graph, and then leverages the LLM to synthesize these into coherent, natural-language responses. Because each answer is grounded in structured evidence, this approach substantially reduces hallucinations commonly observed in standalone LLMs and enhances traceability. When a user submits a query, the system identifies key biomedical entities, maps them onto the UniD$^3$ knowledge graph, and traverses their connections to assemble supporting evidence. As a result, users receive not only direct answers but also the underlying chain of associations (e.g., how a drug is linked to a disease through specific genes or pathways). This transparent, evidence-backed design makes UniD$^3$ Chatbot a practical tool for researchers and clinicians seeking reliable exploration of biomedical relationships.

To qualitatively evaluate UniD$^3$-QA, we present a representative case study comparing its performance with the base Llama 3.3-70B model on a question from the PubMedQA benchmark (**Figure 6**). Given a clinical query about late-night salivary cortisol as a screening test, the base model produces a fluent but non-specific response. While it references relevant concepts such as the dexamethasone suppression test and urinary free cortisol, it fails to ground its answer in specific studies and ultimately hedges with "it is difficult to conclude definitively." This highlights a common limitation of standalone LLMs: they can recall general knowledge but struggle to synthesize evidence from primary research. In contrast, UniD$^3$-QA adopts an evidence-driven approach by combining retrieved literature with structured knowledge graph relationships to produce precise, verifiable answers. As shown in **Figure 6**, the system retrieves three literature passages related to airflow obstruction, eosinophilic inflammation, and the effects of inhaled corticosteroids on dyspnea perception, along with two knowledge graph entries on asthma pathophysiology and treatment mechanisms. By integrating these five evidence sources, UniD$^3$-QA arrives at a concrete conclusion that, inhaled corticosteroids influence dyspnea perception by reducing airway inflammation and improving bronchial hyperresponsiveness, and supports this claim with numbered, traceable citations. This ability to generate specific, evidence-backed responses, rather than vague generalities, underscores a key advantage of UniD$^3$-QA over standalone LLMs, particularly in biomedical domains where factual accuracy and source transparency are critical.

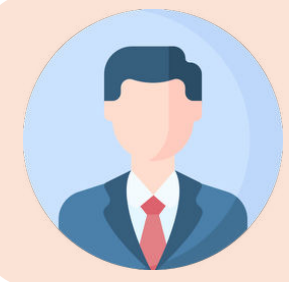

IS LATE-NIGHT SALIVARY CORTISOL A BETTER SCREENING TEST FOR POSSIBLE CORTISOL EXCESS THAN STANDARD SCREENING TESTS IN OBESE PATIENTS WITH TYPE 2 DIABETES?

**UniD$^3$ Answer**

Based on the retrieved evidence, inhaled corticosteroids (ICSs) appear to influence the perception of dyspnea (PD) during bronchoconstriction in asthma patients through multiple mechanisms. Research indicates that baseline airflow obstruction affects a patient's ability to detect further increases in airway resistance [1], and that eosinophilic inflammation of the airway wall can heighten PD [2]. Importantly, ICS treatment has been shown to reduce eosinophilic inflammation, which in turn lessens the perception of dyspnea during induced bronchoconstriction [3]. These findings are consistent with established knowledge of asthma pathophysiology [4] and the recognized role of ICSs in reducing airway inflammation and hyperresponsiveness [5].

Conclusion and References:

The study suggests that inhaled corticosteroids can affect the perception of dyspnea during bronchoconstriction in asthma patients by reducing airway inflammation and improving bronchial hyperresponsiveness. This finding has important implications for the management of asthma symptoms and the use of ICSs in treatment regimens.

References:

1. [VD] Context: Influence of baseline airflow obstruction on the patient's ability to detect any further increase in airway resistance.
2. [VD] Context: Effect of eosinophilic inflammation on the airway and its relation to perception of dyspnea.
3. [VD] Study findings on the effect of ICSs on PD during bronchoconstriction.
4. [KG] Asthma pathophysiology and treatment guidelines.
5. [KG] Role of inhaled corticosteroids in asthma management and their effects on airway inflammation and hyperresponsiveness."

**Reference Answer**

Long: We have shown that eosinophilic inflammation of the airway wall may increase PD and that the association of eosinophil counts with ICSs may result in lessening the PD.

Short: yes

**Base Model Answer**

Late-night salivary cortisol (LNSC) has been proposed as a screening test for cortisol excess. In general, LNSC has shown reasonable sensitivity for detecting Cushing's syndrome. However, in obese patients with type 2 diabetes, the interpretation may be complicated by metabolic factors that influence cortisol levels. Standard screening tests such as the 1-mg overnight dexamethasone suppression test and 24-hour urinary free cortisol remain widely used. Without specific evidence comparing these modalities in this patient population, it is difficult to conclude definitively whether LNSC is superior.

**Figure 6 Comparative case study of UniD³-QA versus base model Llama 3.3-70B on a biomedical question-answering task.** A question from the PubMedQA benchmark is presented alongside the reference answer (top). The base Llama 3.3-70B model (right) generates a plausible but generic response without domain-specific grounding. In contrast, UniD³-QA (left) produces a grounded answer by integrating retrieved literature passages from a vector database which is automatically built during knowledge graph construction via LightRAG and structured entries from the knowledge graph, with source tags (e.g., [VD] for vector database, [KG] for knowledge graph) indicating the provenance of each cited reference.

## DISCUSSION

In this work, we present UniD$^3$, a novel framework that integrates Large Language Models (LLMs) and Knowledge Graph-enhanced Retrieval-Augmented Generation (KG-RAG) to systematically mine, extract, and organize drug-disease relationships from biomedical literature. By employing a dual-stage entity extraction strategy, UniD$^3$ significantly improves the efficiency, accuracy, and scalability of drug-disease knowledge organization and dataset construction. UniD$^3$ addresses the existing gap between unstructured biomedical knowledge and structured, machine-readable knowledge resources. Our main contributions are as follows: To the best of our knowledge, UniD$^3$ is the first framework to leverage LLMs combined with KG-RAG technology to jointly construct biomedical knowledge graphs, generate multi-task Drug-Disease datasets, and deliver an explainable question-answering interface from unstructured biomedical text. A comprehensive set of knowledge graphs and multi-task datasets has been constructed using UniD$^3$, validated by clinicians and cross-referenced with external datasets. The resulting knowledge graphs serve as reusable biomedical resources with broad applicability, and the accompanying explainable chatbot (https://unid3.site/) enables researchers and clinicians to interactively explore drug-disease associations grounded in structured evidence. We further benchmark both general-purpose and biomedical-specific LLMs on the UniD$^3$-generated datasets to assess their performance and applicability in Drug-Disease question answering.

A noteworthy aspect of the UniD$^3$ datasets lies in their overlap with established curated drug repositories such as FDA-approved drug lists and DrugBank. Although the UniD$^3$ datasets contain fewer total drugs than these regulatory-level resources, the intersection across FDA, DrugBank, and the UniD$^3$ datasets is remarkably limited. Rather than indicating incompleteness, this reflects the complementary nature of UniD$^3$. While FDA and DrugBank are inherently biased toward compounds with late-stage clinical validation and regulatory approval, UniD$^3$ systematically captures drug-related relationships from preclinical research, mechanistic studies, and early-stage clinical investigations. Consequently, the UniD$^3$ datasets are enriched for understudied compounds, exploratory therapeutic agents, and mechanistic associations that are not yet represented in regulatory-approved repositories. This complementary coverage suggests that UniD$^3$ may serve as an intermediate knowledge layer connecting hypothesis-driven biomedical discovery with clinically validated evidence, thereby expanding the search space for drug repurposing and translational hypothesis generation. At the same time, the presence of well-established anticancer agents such as Cisplatin, Doxorubicin, Tamoxifen, and EGFR inhibitors among the top-ranked graph nodes demonstrates that UniD$^3$ reliably captures clinically meaningful pharmacological associations rather than noise.

UniD$^3$ offers several important advantages over traditional manually curated or rule-based drug-disease knowledge construction pipelines. By integrating LLM-driven extraction with KG-enhanced retrieval and dual-stage entity consolidation, the framework achieves a strong balance between scalability, efficiency, and data quality, while retaining interpretability through explicit entity-relationship grounding. The resulting datasets demonstrate high agreement with clinician validation and external benchmarks, the constructed knowledge graphs exhibit dense, well-connected structures suitable for downstream reasoning, and the explainable chatbot provides evidence-grounded responses with traceable citations, collectively supporting diverse biomedical applications such as drug repurposing, pharmacovigilance, and mechanistic inference. Nevertheless, several limitations should be acknowledged. Because UniD$^3$ relies on LLMs for entity and relationship extraction, issues such as misclassification, incomplete recognition, and occasional hallucination may still occur, potentially introducing noise into the knowledge graphs and datasets. Moreover, dependence on PubMed literature may introduce publication bias toward positive or well-studied findings, limiting coverage of rare conditions or under-reported therapeutic contexts. Although cross-dataset validation improves generalizability, certain associations remain context-specific and may not directly translate across populations or disease subtypes. At the knowledge-graph level, prioritization of direct Drug-Disease links may underrepresent multi-step biological pathways and indirect mechanistic relationships. Additionally, while KG-RAG grounding improves factual reliability, the LLM-driven extraction process still presents challenges for full transparency and explainability. Finally, as with any automatically generated biomedical resource, careful contextual interpretation remains essential to avoid overstating weak or uncertain associations.

Overall, UniD$^3$ establishes a scalable, cost-effective, and extensible framework for transforming unstructured biomedical literature into structured Drug-Disease knowledge, supporting applications ranging from hypothesis generation to model benchmarking. Future work will focus on extending the breadth and depth of knowledge graphs and datasets through continuous integration of newly published research and regulatory document, incorporating multi-modal biomedical data such as molecular, genetic, and clinical outcome evidence, developing evidence-aware confidence scoring to better distinguish high-confidence therapeutic associations from exploratory research signals, and enhancing the chatbot with multi-turn dialogue capabilities and user feedback mechanisms. In parallel, further refinement of explainability mechanisms and expert-in-the-loop feedback will enhance both transparency and reliability. Through these ongoing developments, UniD$^3$ is expected to evolve into a continuously learning knowledge platform that more tightly links mechanistic discovery with clinically actionable insight through integrated knowledge graphs, datasets, and conversational AI.

## DATA AVAILABILITY

All datasets used in this study are publicly accessible, including the Drug-Disease Matching (DDM) dataset at https://huggingface.co/datasets/Mike2481/UniD3_DDM, the Drug-Target Analysis (DTA) dataset at https://huggingface.co/datasets/Mike2481/UniD3_DTA, and the Drug Effectiveness Assessment (DEA) dataset at https://huggingface.co/datasets/Mike2481/UniD3_DEA.

## CODE AVAILABILITY

The code used to develop the model and generate the results in this study is publicly available under the MIT License at https://github.com/QSong-github/UniD3. In addition, the UniD$^3$ web chatbot (https://unid3.site/) provides interactive access to the UniD$^3$-generated datasets, enabling users to explore and query the resources.

## AUTHOR CONTRIBUTIONS

QW (Conceptualization, Methodology, Formal analysis, Writing - original draft), TL (Result Evaluation, Writing - review & editing), MZ (Data curation, Writing - review & editing, web chatbot), JL (Writing - review & editing), SG (Writing - review & editing), GW (Conceptualization, Writing - review & editing); JS (Conceptualization, Writing - review & editing); QS (Conceptualization, Supervision, Project administration, Formal analysis, Funding acquisition, Writing - review & editing).

## CONFLICTS OF INTEREST

The authors have no competing interests to declare.

## FUNDING

Q. S. is supported by the National Institute of General Medical Sciences of the National Institutes of Health (R35GM151089). G.W. is supported by the National Institute of General Medical Sciences of the National Institutes of Health (1R35GM150460). J.S. is financially supported by the National Library of Medicine of the National Institute of Health (R01LM013771), the National Cancer Institute of the National Institutes of Health (P30CA082709-25S1), the Indiana University Precision Health Initiative, and the Indiana University Melvin and Bren Simon Comprehensive Cancer Center Support Grant from the National Cancer Institute (P30CA 082709). This work partially used through allocation CIS230237[58] from the Advanced Cyberinfrastructure Coordination Ecosystem: Services Support (ACCESS)[59] program, which is supported by National Science Foundation grants #2138259, #2138286, #2138307, #2137603, and #2138296.